\documentclass[12pt]{article}

\usepackage{makecell}
\usepackage[T2A]{fontenc}
\usepackage[utf8]{inputenc}  
\usepackage{csquotes}

\usepackage{newlfont}
\usepackage{supertabular}
\usepackage{amsmath}
\usepackage{amssymb}
\usepackage{amsfonts}
\usepackage{amsthm}
\usepackage{bm}
\usepackage{mathtools}
\usepackage{pb-diagram}
\usepackage{righttag}
\usepackage{eqnarray}
\usepackage{subeqnarray}
\usepackage{algorithmic}
\usepackage{epsfig}
\usepackage{subfigure} 
\usepackage{emlines2}
\usepackage{rotating}
\usepackage{multirow}
\usepackage{graphicx}

\usepackage{tabularx}
\usepackage{parnotes}
\usepackage[backend=biber,style=numeric,sorting=none]{biblatex}
\usepackage{xurl}
\usepackage{hyperref}
\usepackage[english]{babel}

\DeclareFieldFormat{labelnumber}{#1} 

\usepackage{color}

\usepackage{booktabs}
\usepackage{url}
\usepackage{float}

\usepackage{fancybox}

\theoremstyle{plain}

\theoremstyle{definition}
\newtheorem{example}{Пример}

\newtheoremstyle{break}
  {9pt}
  {9pt}
  {\itshape}
  {}
  {\bfseries}
  {.}
  {\newline}
  {}

\theoremstyle{break}

\makeatletter

\renewcommand{\section}{\@startsection{section}{1}%
             {\parindent}{3.5ex plus 1ex minus .2ex}%
             {2.3ex plus.2ex}{\normalsize\bf}}

\renewcommand{\subsection}{\@startsection{subsection}{2}%
             {\parindent}{3.5ex plus 1ex minus .2ex}%
             {2.3ex plus.2ex}{\normalsize\bf}}

\renewcommand{\subsubsection}{\@startsection{subsubsection}{3}%
             {\parindent}{3.5ex plus 1ex minus .2ex}%
             {2.3ex plus.2ex}{\normalsize\bf}}

\renewcommand{\paragraph}{\@startsection{paragraph}{4}%
             {\parindent}{3.5ex plus 1ex minus .2ex}%
             {2.3ex plus.2ex}{\normalsize\bf}}

\renewcommand{\@biblabel}[1]{#1\hfill}

\makeatother

\graphicspath{{Picts/}}

\newcounter{partnumber}

\makeatletter
\@addtoreset{equation}{section}
\renewcommand{\theequation}{\arabic{equation}}

\@addtoreset{table}{section}

\makeatother

\newcounter{fragm}
\newcounter{subfragm}[fragm]

\newcounter{myremark}[section]

\newcounter{myalgorithm}[section]

\usepackage{float}
\usepackage{tabularx}

\usepackage{xcolor}
\usepackage{nicematrix}

\newcolumntype{C}{>{\centering\arraybackslash}X}  
\newcolumntype{P}[1]{>{\centering\arraybackslash}p{#1}} 

\usepackage{caption}
\newlength{\DepthReference}
\newlength{\HeightReference}
\newlength{\Width}%

\newcommand{\MyColorBox}[2][red]%
{%
    \settowidth{\Width}{#2}%
    \colorbox{#1}%
    {%
        \raisebox{-\DepthReference}%
        {%
                \parbox[b][\HeightReference+\DepthReference][c]{\Width}{\centering#2}%
        }%
    }%
}
\newcommand{\MyTBox}[2]%
{%
    \settowidth{\Width}{#1}%
    \fbox%
    {%
        \raisebox{-\DepthReference}%
        {%
                \parbox[b][\HeightReference+\DepthReference][c]{\Width}{\centering#1}%
        }%
    }%
}

\usepackage{xcolor}

\begin{document}

\begin{center}
\textbf{RCDINO: Enhancing Radar-Camera 3D Object Detection \\ with DINOv2 Semantic Features}
\end{center}

\begin{center}
Matykina Olga$^{1*}$, Yudin Dmitry$^{1,2}$\\
$^{1}$Moscow Institute of Physics and Technology, Moscow, Russia \\
$^{2}$AIRI, Moscow, Russia \\
$^{*}$e-mail: matykina.ov@phystech.edu \\
\end{center}

\textbf{Abstract.}
Three-dimensional object detection is essential for autonomous driving and robotics, relying on effective fusion of multimodal data from cameras and radar. This work proposes RCDINO, a multimodal transformer-based model that enhances visual backbone features by fusing them with semantically rich representations from the pretrained DINOv2 foundation model. This approach enriches visual representations and improves the model's detection performance while preserving compatibility with the baseline architecture. Experiments on the nuScenes dataset demonstrate that RCDINO achieves state-of-the-art performance among radar-camera models, with 56.4 NDS and 48.1 mAP. Our implementation is available at \url{https://github.com/OlgaMatykina/RCDINO}.


\bigskip
\section{Introduction}
\label{sec:intro}

Three-dimensional perception of the surrounding environment plays a key role in autonomous driving, robotics, and intelligent transportation systems. One of the central tasks in this field is 3D object detection, which enables safe and accurate interaction between an agent and the real world. Multimodal 3D detection, where data from various sensors such as cameras, radars, and LiDARs are combined for robust and precise object detection, is particularly important.

Cameras provide high-resolution images rich in texture and color information, but they lack direct depth measurements. In contrast, radar offers reliable distance and velocity information even under adverse weather conditions \cite{zhang2023perception}, but suffers from low spatial resolution and a lack of semantic detail. Effectively fusing these complementary modalities remains a significant and ongoing challenge.

In recent years, numerous approaches have been proposed for radar-camera 3D detection. Some perform feature fusion in three-dimensional space using the bird’s-eye view (BEV) representation, which enables the unification of data from different modalities on a common plane. Other methods adopt query-based architectures, where object queries are progressively refined to localize targets. Despite these advancements, existing solutions often suffer from limited semantic richness of visual features, which hampers the detection of distant objects - an essential capability for reliable autonomous perception.

In this paper, we propose Radar-Camera DINO (RCDINO), a novel radar-camera 3D detection framework that enhances the standard visual backbone with semantically rich features extracted by the pretrained DINOv2 model~\cite{dinov2}. To effectively incorporate these features, we introduce a lightweight adapter module that fuses DINOv2 representations into the existing pipeline.

Our design is inspired by recent advances in query-based architectures such as RCTrans~\cite{li_rctrans_2024}. In the ablation study, we investigate the impact of decoder components from the baseline model and find that the two-stage decoder improves detection performance even when one modality provides less informative features. This highlights the significant role of the decoder architecture in effective multimodal fusion.

We validate our approach on the nuScenes dataset~\cite{caesar2020nuscenes} and on real-world radar-camera data collected by our team, demonstrating strong detection performance and robustness in practical settings. Our main contributions are:

\begin{itemize}
    \item Introducing RCDINO, a radar-camera 3D detection model that enriches visual features with pretrained DINOv2 representations via a lightweight adapter.
    \item Providing an ablation study that analyzes the benefits of the two-stage decoder design independent of feature informativeness.
    \item Validating the approach on both public benchmarks and real-world collected data, confirming generalization and practical utility.
    \item Releasing our implementation to foster further research and reproducibility: \url{https://github.com/OlgaMatykina/RCDINO}.
\end{itemize}

\section{Related Work}
\label{sec:related_works}
\subsection{BEV-based 3D object detection}
The most common approach to 3D object detection is based on the Bird’s Eye View (BEV) representation, where data from various sensors is projected into a unified coordinate system. A major challenge in constructing BEV from camera images lies in inaccurate depth estimation, which BEVDepth~\cite{li2023bevdepth} addresses through the introduction of a learnable depth module. To improve object velocity prediction and reduce missed detections, many works incorporate temporal information. BEVFormer~\cite{li2022bevformer} was the first to introduce sequential temporal modeling into multi-camera 3D object detection. SOLOFusion~\cite{park2022solofusion} further demonstrates that long-term and short-term temporal fusion complement each other well, proposing a corresponding dual-stream fusion approach. Another method that employs BEV and temporal fusion to enable feature map propagation is FMFNet~\cite{murhij2022fmfnet}.

BEV can also serve as a unified representation space for feature fusion across modalities. BEVFusion~\cite{liu2023bevfusion} was one of the first methods to use BEV as a common representation space for multimodal fusion. The BEV representation allows for explicit alignment of data from different sources, making it highly suitable for cross-modal integration. Another work, BEVCar~\cite{schramm2024bevcar}, proposes a unified approach for object and map segmentation in BEV space. Its key novelty lies in initially learning point-wise encoding from raw radar data, which is then used to efficiently initialize the projection of image features into the BEV space. The backbone used is the foundation model DINOv2\cite{dinov2} with an adapter - an idea that inspired our method.

Despite the advantages of using a unified BEV space, misalignment errors between different BEV representations can degrade fusion quality. Some approaches tackle this with attention mechanisms. For instance, RCBEVDet~\cite{lin_rcbevdet_2024} transforms radar points into BEV features via RadarBEVNet and fuses them with image features using a cross-attention module. Similarly, CRN~\cite{kim_crn_2023} applies deformable cross-modal attention to align image and radar features in the BEV space.

\subsection{Beyond BEV methods}
There are also alternative methods outside the BEV paradigm. A notable image-only approach is ImVoxelNet\cite{rukhovich2022imvoxelnet}, which features a convolutional architecture capable of handling both single and multi-view images from differently positioned cameras. It performs well in both indoor and outdoor settings, differing only in detection heads depending on the domain. DAGM-Mono~\cite{murhij2024dagm} tackles monocular 3D pose estimation and shape reconstruction using a deformable attention-guided framework. By introducing Chamfer-based losses and leveraging inter-object as well as scene context, it achieves state-of-the-art results on ApolloCar3D and improves the accuracy of existing monocular 3D detectors. MVFusion\cite{wu_mvfusion_2023} introduces the semantically aligned radar encoder SARE and employs transformers to build robust features. SpaRC\cite{wolters_sparc_2024} uses frustum-based fusion and local self-attention to enhance localization accuracy. HyDRa\cite{wolters_unleashing_2024} applies a Height Association Transformer to improve depth estimation using radar features. RVCDet~\cite{murhij2022rethinking} addresses real-time performance and reliability in LiDAR-based 3D detection by introducing a fast dynamic voxelizer compatible with pillar-based models and a lightweight sub-head for filtering false positives.

\subsection{Query-based 3D object detection}
Query-based methods represent another major paradigm in 3D object detection. In these approaches, object queries interact with input tokens via attention mechanisms to refine predictions. DETR3D\cite{wang2022detr3d} introduces a set of 3D reference points, each associated with an object query. Subsequent research has focused on more efficient extraction of image features for these queries. PETR\cite{liu2022petr} proposes positional encoding of image features in LiDAR coordinates. CAPE\cite{xiong2023cape} constructs positional encoding in camera coordinates to reduce the influence of camera extrinsic changes. StreamPETR\cite{wang2023streampetr} proposes a query-based long-term temporal fusion algorithm that carries historical object information into the current frame.

Modern methods increasingly combine the strengths of both BEV and query-based paradigms. RaCFormer\cite{chu2024racformer}, for example, extracts features relevant to object queries both from BEV space and directly from images, mitigating depth estimation errors by learning depth maps from radar and image features. It also introduces adaptive query initialization in polar coordinates and leverages Doppler-based motion cues in BEV features to enhance temporal modeling. RCTrans\cite{li_rctrans_2024} likewise implements a query-based scheme but focuses on enriching radar features using a Radar Dense Encoder, which is then fused with image features at the token level. A sequential decoder is employed to iteratively refine object positions. 

\section{Method}
\label{sec:method}

\begin{figure}[h]
    \centering
    \includegraphics[width=1\linewidth]{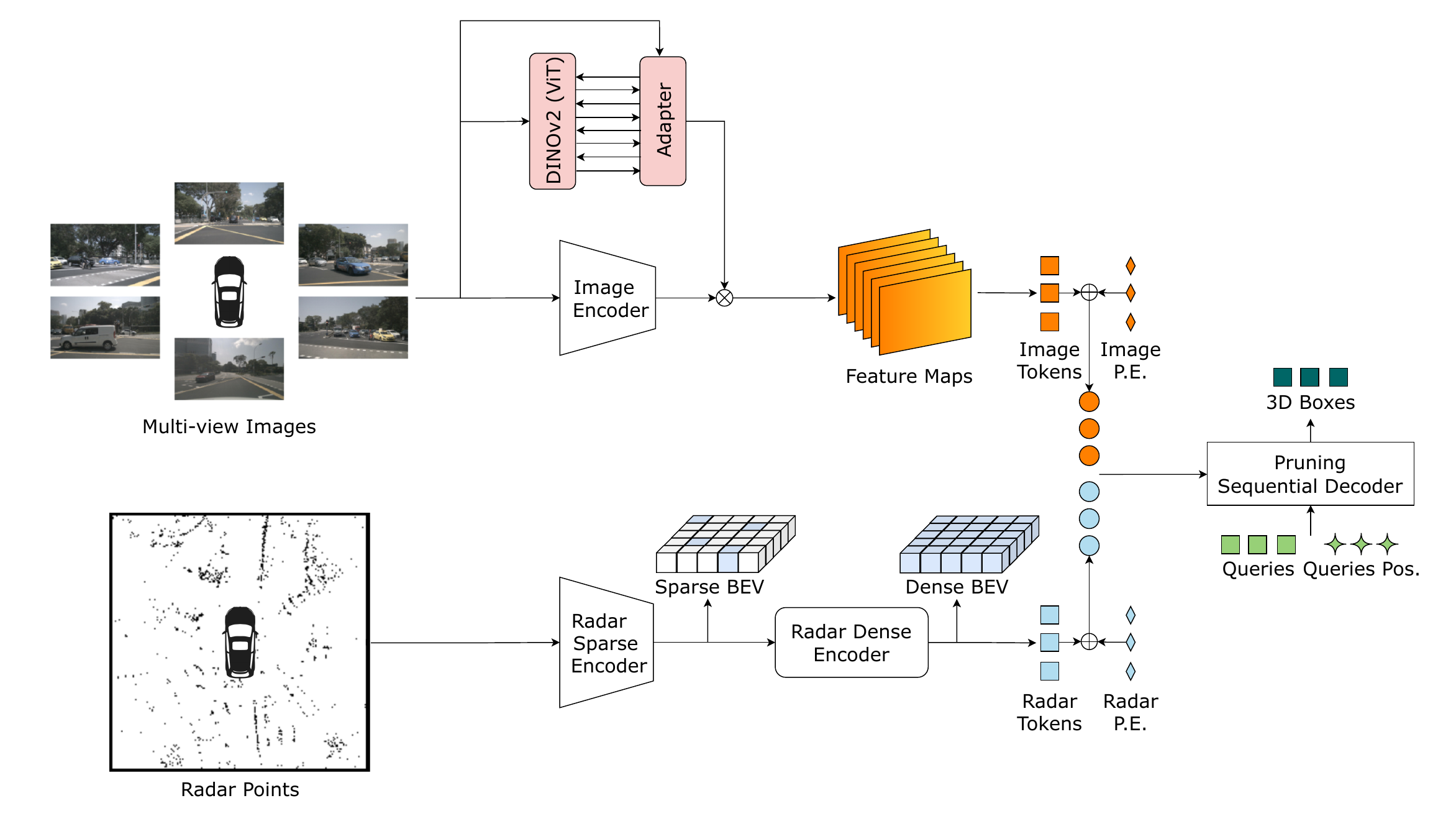}
    \caption{Architecture of the RCDINO model. The key novelty is the DINOv2 adapter, which integrates semantically rich features into the visual backbone. 
The camera and radar branches can be individually disabled; the impact of each branch is analyzed in the Ablation Study (Section~\ref{sec:ablation}).}
    \label{fig:rctrans_dinov2}
\end{figure}

The proposed model, \textbf{RCDINO} (Radar-Camera DINO), is a multimodal 3D object detector with a transformer-based decoder that iteratively refines object queries using both camera and radar features. As shown in Figure~\ref{fig:rctrans_dinov2}, the architecture consists of five main components: a visual encoder, a DINOv2 adapter, a sparse radar encoder, a dense radar encoder, and a sequential decoder.

\subsection{Visual Encoder}
For each image from the multi-camera setup, visual features are extracted using a standard convolutional or transformer-based backbone, such as ResNet~\cite{he2016deep} or Vision Transformer (ViT)~\cite{dosovitskiy2020image}. These features are compressed representations that are later fused with DINOv2 features for enhanced scene understanding.

\subsection{DINOv2 Adapter}

To integrate semantically rich features from a pretrained DINOv2 model into our detection pipeline, we employ a lightweight adapter module that interacts with the DINOv2 backbone in two directions - injection and extraction.

The adapter operates as follows:

\begin{itemize}
    \item First, input image features from the standard convolutional backbone (e.g., ResNet18) are processed by a \textbf{convolutional feature extractor} to produce multi-scale spatial features aligned with DINOv2's internal resolution.
    \item These features are then injected into the intermediate layers of DINOv2 using a module we call the \textbf{injector}. The injector performs deformable attention and normalization to fuse the external features with the internal DINOv2 representations. The influence of the injected features is modulated by a learnable scalar gate.
    \item DINOv2 then performs inference as usual, allowing the injected features to propagate through its transformer layers and be refined in the semantic context of large-scale pretraining.
    \item After inference, the modified features are retrieved using the \textbf{extractor}, which applies convolutional feed-forward layers with spatial feedback to extract task-specific representations from the modified DINOv2 features.
\end{itemize}

The extracted features are interpolated and projected to match the dimensions of the main visual backbone. A \textbf{learnable fusion layer} then combines the original backbone features with the extracted DINOv2 features.

This adapter structure allows DINOv2 to serve as a semantic enhancer for visual perception: it refines external task-specific features through interaction with its pretrained knowledge, and then returns enriched representations to the detection network. Importantly, the adapter does not modify DINOv2 weights unless explicitly fine-tuned, preserving the benefits of large-scale pretraining while remaining computationally efficient.

\subsection{Radar Sparse Encoder}
Radar points are processed using a pillar-based approach inspired by Futr3D~\cite{chen2023futr3d}. The radar input $P \in \mathbb{R}^{N_r \times 5}$ includes 3D coordinates, compensated velocity, and time offset. A multi-layer perceptron (MLP) extracts point-wise features that are aggregated into a BEV grid. This component is structurally similar to the baseline radar encoder.

\subsection{Radar Dense Encoder}
To mitigate the sparsity of radar signals, a dense encoder inspired by U-Net~\cite{ronneberger2015u} is employed. A dense BEV radar map $B \in \mathbb{R}^{H_r \times W_r \times C_r}$ is constructed, from which multi-resolution features are extracted: $B_d = \{B_i \in \mathbb{R}^{\frac{H_r}{2^i} \times \frac{W_r}{2^i} \times C_i}, i = 1, 2, 3\}$. Self-attention is applied at the lowest resolution ($B_3$), producing an adaptively enhanced feature map $B_f$, augmented with 2D positional encoding. The final output is upsampled and skip-connected with intermediate $B_d$ layers, in line with U-Net-style designs.

\subsection{Sequential Transformer Decoder}
Multimodal data is fused using a lightweight sequential decoder, which processes camera and radar features in a cascaded fashion rather than through a unified block. This structure simplifies feature integration and reduces computational overhead while supporting joint spatial reasoning.

The decoding sequence begins with object queries that include positional embeddings. These queries are first refined using camera-derived BEV features, followed by refinement with radar range-view (RV) features. Such staged attention allows complementary spatial and semantic information from each modality to be utilized effectively.

Temporal consistency is maintained by incorporating memory features from earlier frames. Each stage attends to these via deformable attention, using learned positional encodings. This approach supports tracking and improves robustness in dynamic or occluded scenes.

At each stage, object queries are updated through learned transformations. Reference point refinement is performed using:
\[
R^{n+1} = R^n + \Delta R,
\]
where $R^n$ denotes the current reference position, and $\Delta R$ the predicted offset. Updated queries are processed through classification and regression heads to generate object categories and 3D bounding boxes.

The final output includes predicted class labels, 3D boxes, and intermediate query states. The decoder's modular and lightweight structure supports scalable multimodal and temporal reasoning.

\section{Experiments}

\subsection{Dataset and Metrics}

As in the original work \cite{li_rctrans_2024}, experiments were conducted on the widely used autonomous driving dataset with 3D object detection annotations - nuScenes \cite{caesar2020nuscenes}. This dataset contains 1000 scenes, split into three parts: 700 for training, 150 for validation, and 150 for testing. Each frame in nuScenes includes 6 camera images and 5 radar point clouds, covering a full $360^\circ$. The dataset contains approximately 1.4 million labeled 3D bounding boxes across ten classes: car, truck, bus, trailer, construction vehicle, pedestrian, motorcycle, bicycle, traffic cone, and barrier. Some classes are also annotated with additional attributes that describe the object's state, such as whether it is stationary or moving.

The primary evaluation metrics for 3D detection are the nuScenes Detection Score (NDS) and mean Average Precision (mAP). NDS is a weighted sum of mAP and several other metrics defined by the dataset authors, including: Average Translation Error (ATE) (meters), Average Scale Error (ASE) (1 - 3D IoU), Average Orientation Error (AOE) (radians), Average Velocity Error (AVE) (m/s), Average Attribute Error (AAE) (1-accuracy).

\subsection{Implementation Details}

As in the original RCTrans training procedure \cite{li_rctrans_2024}, information from 4 previous frames is accumulated to make predictions for the current frame.

The number of decoder layers in the transformer is set to 6 during training and 3 during inference. The number of queries, memory queue size, and the number of propagated queries are set to 900, 512, and 128, respectively.

Radar points are accumulated from 6 previous frames, following CRAFT \cite{kim2023a}. The bird’s-eye view (BEV) radar feature map is set to a resolution of $128 \times 128$.

The DINOv2 model \cite{dinov2} was used with an adapter. Weights from the pretrained DINOv2-small model were utilized, with a patch size of $14 \times 14$. To accommodate this model, the input image size was resized to $224 \times 448$ using bilinear interpolation. Features were extracted from 4 layers of DINOv2 and fused using interpolation and convolution. The resulting features were then interpolated to a size of $16 \times 44$ and combined with ResNet features using a learnable weighting coefficient.

The network was trained for 16 epochs, starting from pretrained weights of the original model, with a batch size of 12 and gradient accumulation over 3 steps on 2 NVIDIA A100 GPUs. The speed was evaluated on a single NVIDIA RTX3090 GPU. Optimization was performed using AdamW with a weight decay of $10^{-2}$. The learning rate was scheduled using a fixed step strategy: $1 \times 10^{-6}$ for the first 11 epochs and $1 \times 10^{-7}$ for the final 5 epochs.

\subsection{Results}

\begin{figure}
\centering
\includegraphics[width=1\linewidth]{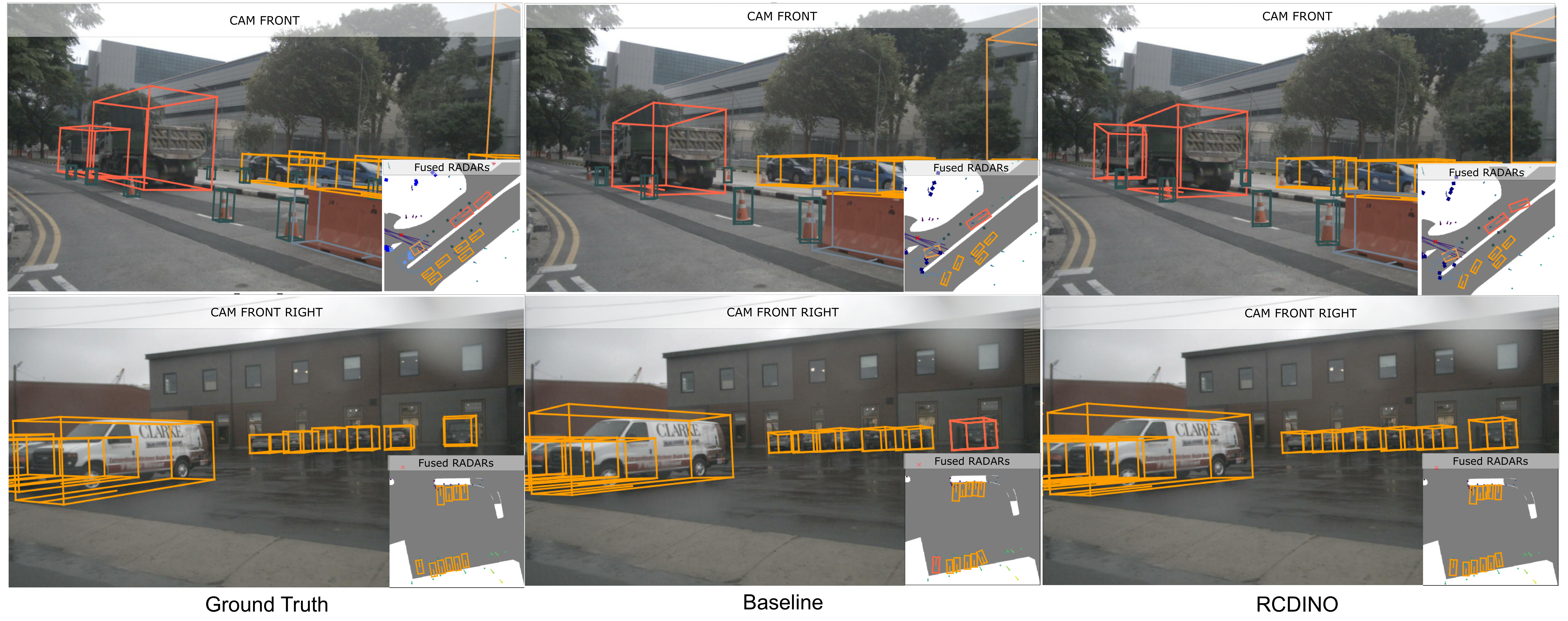}
\caption{Qualitative 3D object detection results on the nuScenes validation set. 
The \textit{top row} demonstrates that enriching visual features enables accurate detection of distant objects, while the \textit{bottom row} shows correct classification of poorly visible objects.}
\label{fig:results_nuscenes}
\end{figure}

The experiment demonstrated that integrating the pretrained DINOv2 model with the RCTrans architecture improves 3D object detection performance on the nuScenes dataset \cite{caesar2020nuscenes}. We observe an increase of $0.4\%$ in NDS and $0.7\%$ in mAP compared to the baseline RCTrans model. Table~\ref{tab:3d_detection_comparison} shows the performance comparison of different 3D detection models on the nuScenes validation set. While RCDINO achieves a modest improvement of +0.4 NDS and +0.7 mAP over the baseline RCTrans, it also introduces an increase in inference latency - from 48.9ms to 85.2ms per frame. Despite nearly doubling inference time, the final latency remains acceptable for many robotic systems and is particularly applicable for low-speed navigation or offline annotation tasks. Moreover, accelerating inference was not the objective of this work. The goal was to assess the potential benefit of incorporating foundation model features into radar-camera fusion architectures with minimal pipeline modification.
Figure~\ref{fig:results_nuscenes} presents qualitative results of 3D object detection on the nuScenes validation set. For illustration purposes, only one camera image is shown per scene, as it reveals the most visible differences.

\begin{table}[ht]
\caption{Comparison of 3D object detection results on the nuScenes validation set. C and R denote camera and radar, respectively. * - inference time is taken from \cite{li_rctrans_2024}.}
\label{tab:3d_detection_comparison}
\centering\medskip
\begin{tabular}{lccccc}
\hline
Method & Input & \makecell{Image \\ Backbone} & NDS (\%) $\uparrow$ & mAP (\%) $\uparrow$ & \makecell{Inference \\ time (ms)}  $\downarrow$  \\
\hline
CRN \cite{kim_crn_2023} & RC & ResNet18 & 54.3 & 44.8 & 35.8* \\
RCBEVDet \cite{lin_rcbevdet_2024} & RC & ResNet18 & 54.8 & 42.9 & \textbf{35.3}* \\
RobuRCDet \cite{yue2025roburcdet} & RC & ResNet18 & 55.0 & 45.5 & 74.8 \\
RCTrans \cite{li_rctrans_2024} & RC & ResNet18 & 56.0 & 47.4 & 48.9 \\
\textbf{RCDINO} & RC & ResNet18 & \textbf{56.4} & \textbf{48.1}  & 85.2 \\
\hline
BEVDepth \cite{li2023bevdepth} & C & ResNet50 & 47.5 & 35.1 & 86.2* \\
BEVPoolv2 \cite{huang2022bevpoolv2} & C & ResNet50 & 52.6 & 40.6 & 60.2* \\
SOLOFusion \cite{park2022solofusion} & C & ResNet50 & 53.4 & 42.7 & 87.7* \\
StreamPETR \cite{wang2023streampetr} & C & ResNet50 & 54.0 & 43.2 & 36.9* \\
CRN \cite{kim_crn_2023} & RC & ResNet50 & 56.0 & 49.0 & 49.0* \\
RCBEVDet \cite{lin_rcbevdet_2024} & RC & ResNet50 & 56.8 & 45.3 & 47.0* \\
RobuRCDet \cite{yue2025roburcdet} & RC & ResNet50 & 56.7 & 51.2 & 113.6 \\
RCTrans \cite{li_rctrans_2024} & RC & ResNet50 & 58.6 & 50.9 & 84.1 \\
\textbf{RCDINO} & RC & ResNet50 & \textbf{59.0} & \textbf{51.4} & 156.9 \\
\hline
\end{tabular}
\end{table}


To better understand the impact of DINOv2 features on detection performance, we report per-class AP for both RCTrans and RCDINO in Table~\ref{tab:perclass_ap}. The results show that RCDINO achieves improvements in several categories, including \textit{truck}, \textit{bus}, \textit{construction vehicle}, \textit{pedestrian}, \textit{bicycle}, and \textit{barrier}. These object types are often semantically complex or visually ambiguous, suggesting that DINOv2's high-level semantic representations can improve detection in challenging cases. In a few dominant classes, such as \textit{car} and \textit{motorcycle}, the baseline achieves slightly higher AP.

\begin{table}[ht]
\centering
\caption{Per-class AP (\%) for RCTrans and RCDINO on the nuScenes validation set. Bold indicates best performance.}
\label{tab:perclass_ap}
\begin{tabular}{l|cccccccccc}
\toprule
\textbf{Model} & car & truck & bus & trailer & const. veh. & ped. & moto & bicycle & cone & barrier \\
\midrule
RCTrans & \textbf{72.9} & 41.3 & 44.3 & \textbf{20.9} & 18.6 & 55.4 & \textbf{49.8} & 44.3 & \textbf{67.1} & 60.9 \\
RCDINO & 72.4 & \textbf{43.2} & \textbf{45.2} & 20.2 & \textbf{19.1} & \textbf{55.6} & 49.3 & \textbf{46.1} & 67.0 & \textbf{61.9} \\
\bottomrule
\end{tabular}
\end{table}

\section{Ablation Study}
\label{sec:ablation}
\subsection{Experiments with the Transformer Decoder}

In addition to modifying the encoder part of the RCTrans architecture \cite{li_rctrans_2024}, we conducted experiments with the transformer decoder and the overall network architecture to investigate the influence of different modalities on 3D object detection performance. The authors proposed an unusual design of the transformer detection head, which includes a sequential refinement of the object query positional embeddings. Based on reference points of the queries, both 2D and 3D positional embeddings are precomputed.

First, 2D positional embeddings of the queries are passed into the transformer layer, which interact via cross-attention with the 2D radar embeddings (in the coordinates of the BEV feature map), updating the queries. Then, in the next transformer layer, 3D positional embeddings of the queries are used and are similarly transformed through cross-attention with the image positional embeddings. The outputs of these two decoder layers update the reference points of the queries, and the whole procedure is repeated. In this way, the query positions are refined iteratively.

However, the authors \cite{li_rctrans_2024} did not investigate how the absence of one of the positional embedding refinement components affects prediction quality. This section presents such experiments and corresponding conclusions.

First, we conducted experiments where one of the modalities was removed from the entire network architecture. Removing either the camera or radar branch naturally led to a decrease in prediction performance. Training was conducted using initialization from the pretrained full model, which helped reduce training time. When an entire sensor modality branch was removed, the corresponding query positional embedding - either 2D or 3D - was also excluded from the decoder. To preserve network depth, the missing component layer was replaced with a layer using the remaining component. As a result, the network consisted of multiple identical transformer layers.

The significant drop in detection quality when one of the modalities was removed emphasizes the importance of both components of query positional encoding in the transformer head. Therefore, we conducted an additional experiment attempting to preserve the query token architecture in its original form, despite the absence of data from one of the sensors.

To meet this condition, the features from the missing sensor in the detection head were replaced with a tensor of ones. As in the previous experiments, the model was initialized with pretrained weights of the full RCTrans model with a ResNet18 backbone. This setup simulates a scenario where one of the sensors fails to provide data.

In the case of radar deactivation, the BEV feature map was replaced with a tensor of ones. If the camera was disabled, the image features from the backbone and neck were also replaced with a tensor of ones matching the expected feature tensor dimensions.

An interesting observation was made: disabling one modality in this way caused less performance degradation for the image-only model than completely removing the radar branch (and thus the 2D positional embeddings). This suggests that preserving the head structure and query positional embeddings - even without informative features - gives the model some advantage. The results of these experiments are shown in Table~\ref{tab:decoder-results}.


\begin{table}[h!]
\centering
\caption{Results of experiments on disabling modalities and components of the transformer decoder. C and R denote camera and radar, respectively.}
\label{tab:decoder-results}
\begin{tabular}{ccccccccc}
\hline
Input & Stages & mAP $\uparrow$ & mATE $\downarrow$ & mASE $\downarrow$ & mAOE $\downarrow$ & mAVE $\downarrow$ & mAAE $\downarrow$ & NDS $\uparrow$ \\
\hline
R & One & 0.0 & 111.5 & 63.4 & 71.1 & 92.8 & 58.3 & 11.5 \\
R & Two & 1.3 & 98.5 & 61.2 & 66.0 & 80.2 & 55.7 & 14.5 \\
C & One & 23.0 & 89.2 & 28.7 & 74.0 & 99.1 & 26.2 & 29.8 \\
C & Two & 30.1 & 79.7 & 28.4 & 63.7 & 51.0 & 25.7 & 40.2 \\
RC & Two & 47.4 & 54.0 & 27.4 & 55.7 & 20.8 & 19.0 & 56.0 \\
\hline
\end{tabular}
\end{table}

\begin{figure}[h!]
\centering
\includegraphics[width=1\linewidth]{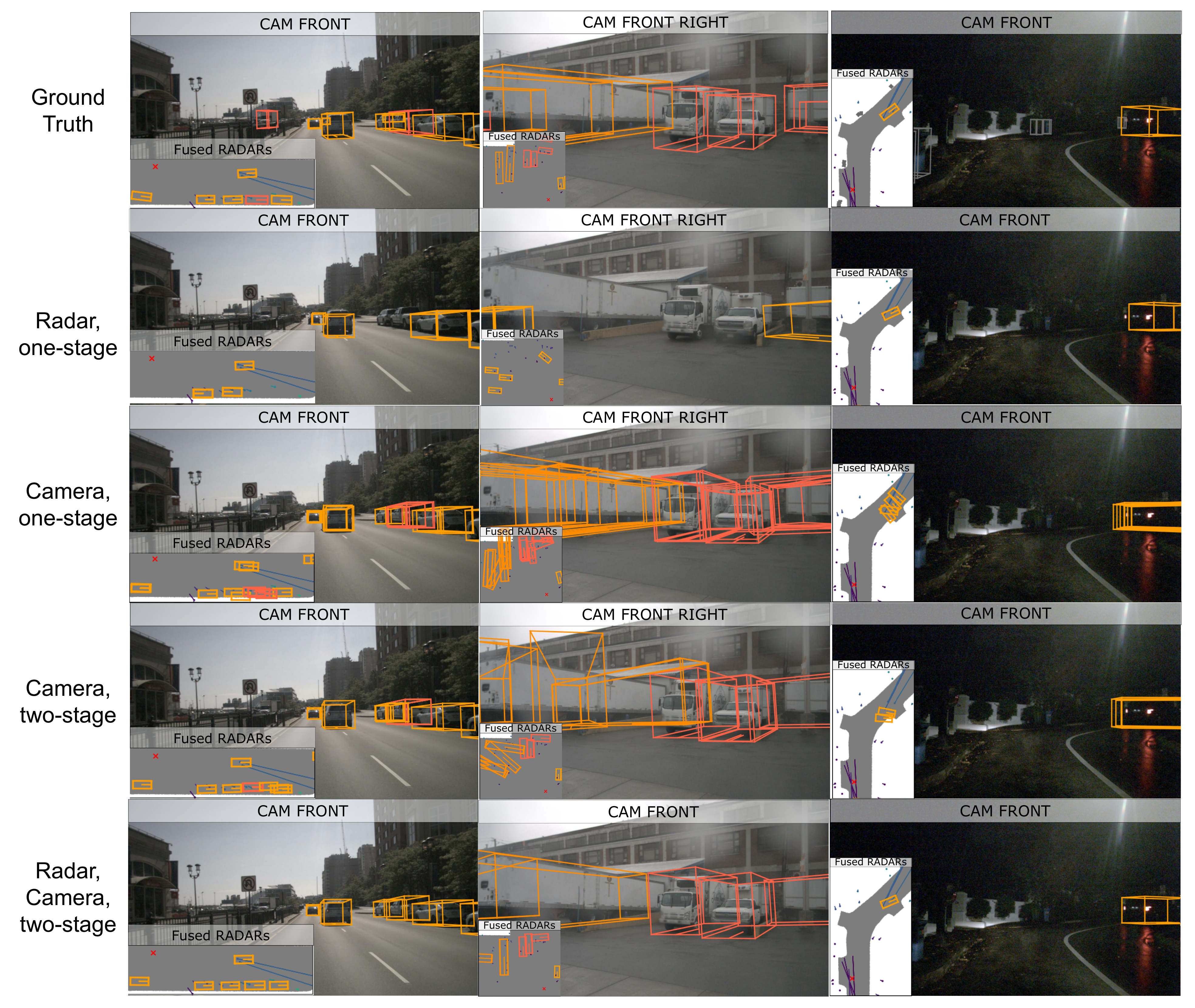}
\caption{Analysis of the influence of radar and camera modalities and decoder components on 3D detection performance on the nuScenes validation set. The \textit{second row} shows that removing the camera modality degrades detection quality. \textit{The third} and \textit{fourth rows} demonstrate that using a two-stage decoder even without informative radar features reduces the model's uncertainty in 3D bbox localization.}
\label{fig:decoder}
\end{figure}

Figure~\ref{fig:decoder} shows qualitative results of 3D object detection on the nuScenes validation set obtained with different radar modality ablation strategies. It can be seen that without the 2D positional embeddings of the queries (i.e., when the radar branch is removed), the model predicts several positions for the same object with high confidence but different depths. This is especially noticeable in the bird’s-eye view visualization. At the same time, preserving the 2D positional embeddings of the queries (i.e., replacing the radar features with a tensor of ones) allows the model to predict object positions more accurately - though still with less precision than the full model. This indicates that preserving the transformer head structure and query positional embeddings - even without meaningful features - gives the model some advantage. 

Although replacing the features from a missing sensor with a tensor of ones may appear overly simplistic, this approach was intentionally chosen to simulate complete sensor failure without introducing additional learnable components or assumptions. One might suggest replacing missing features with average activations computed over the training set. However, this would require accumulating feature statistics for each architecture and training setup, which was not the goal of these ablation experiments. Our objective was to demonstrate that even with minimally informative features (i.e., unit tensors), preserving the structure of the transformer decoder - including the positional embeddings and two-stage refinement - yields significantly better results than removing an entire modality branch. In other words, maintaining the decoder’s structure provides a form of architectural prior that enables the network to better handle missing inputs.

\subsection{DINOv2-only Backbone.}
We also evaluated whether the pretrained DINOv2-small model could serve as a standalone image encoder in place of the standard ResNet18 backbone. In this setup, the DINOv2 features were extracted and directly passed to the detection head, with no ResNet-based encoder present. The model was trained for 22 epochs using the same training schedule and hyperparameters as in other experiments, starting from pretrained RCTrans weights. The resulting performance was extremely poor: the model failed to learn meaningful object detections, achieving 0.0\% mAP and only 6\% NDS. Other metrics were similarly degraded (mATE = 1.076, mASE = 0.839, mAOE = 1.082, mAVE = 0.872, mAAE = 0.692). These results confirm that, despite its semantic richness, DINOv2 alone lacks the spatial resolution and architectural inductive bias required for accurate 3D object detection. Our findings support the use of DINOv2 features as a complement to, rather than a replacement for, a trainable visual backbone.

\subsection{Inference on the Collected Dataset}

To evaluate the generalization ability of the RCDINO model, we used the VegaFull dataset \cite{matykina2025hybrid}, collected during field tests. 
The VegaFull dataset includes synchronized data from a front-facing camera and a 4D radar collected in three geographically and contextually diverse Russian regions: Dolgoprudny, Chekhov, and Aldan. The total volume of data is about 2 TB, comprising 1280x720 RGB images captured under various weather and terrain conditions.

\begin{itemize}
  \item \textbf{VegaDolgoprudny} was collected during off-road driving to capture road defects such as potholes and puddles, providing rich variability in road edge geometry and surface features.
  \item \textbf{VegaChekhov} contains scenes from an industrial area, with construction machinery, dirt mounds, fences, and obstacles specific to structured off-road driving.
  \item \textbf{VegaSeligdar} was recorded at a snow-covered quarry in winter conditions, including low-light and sub-zero temperature scenes, with moving machinery and human workers in safety gear.
\end{itemize}

Thanks to its environmental variability, VegaFull is suitable for evaluating the robustness of radar-camera fusion models in realistic scenarios and adverse conditions.


To assess the generalization of the RCDINO model on the collected VegaFull dataset, an inference procedure was conducted. Due to the difference in sensor configuration - specifically, the presence of only one camera and radar - the model was adapted to these new conditions. The RCDINO model was fine-tuned on the nuScenes dataset using only the front-facing camera and radar. This allowed the model to adjust to the new setup and perform 3D object detection on the collected dataset without attempting to predict objects from directions for which no sensor data were available.

\begin{figure}
\centering
\includegraphics[width=1\linewidth]{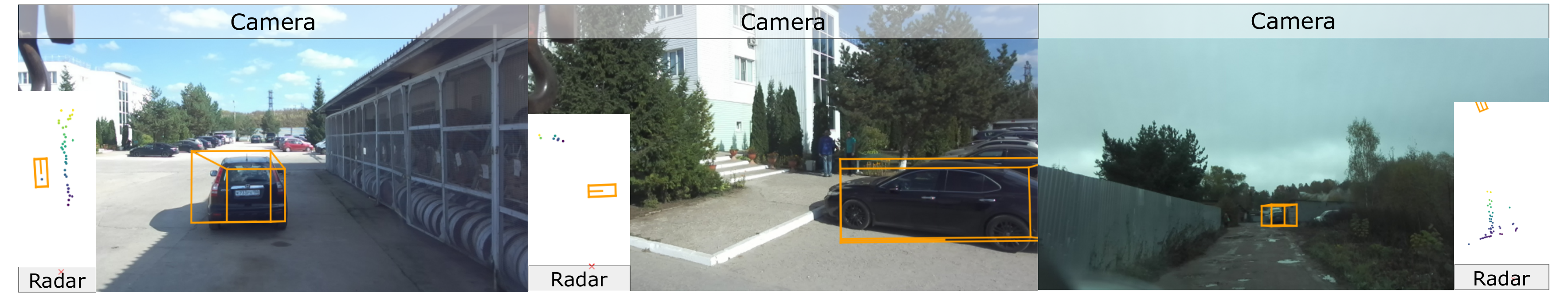}
\caption{3D object detection results of the RCDINO model on the collected VegaFull dataset}
\label{fig:vega_results}
\end{figure}

Figure~\ref{fig:vega_results} shows the 3D object detection results on the VegaFull dataset. The model successfully detects objects, demonstrating its generalization ability and robustness to new data. However, during the collection of the VegaFull dataset, no information about the vehicle’s position in space was recorded, which is critical for temporal alignment of predictions and radar point cloud accumulation. In the absence of such information, only the radar point cloud from the current frame was used at each moment in time. Temporal alignment of predictions was carried out using artificial odometry data from the nuScenes dataset, which could not provide accurate object positioning in space.

To improve prediction quality in the future, we plan to implement a visual odometry algorithm that will use camera data to estimate the vehicle’s position in space. This will allow, first, the accumulation of radar point clouds from previous frames, and second, the use of vehicle motion information to incorporate previous predictions and improve the accuracy of current frame predictions.

\section{Conclusion}

In this work, we presented RCDINO, an enhanced version of the RCTrans architecture for radar-camera 3D object detection. Our approach integrates semantically rich visual features from the pretrained DINOv2 foundation model into the RCTrans pipeline via a lightweight adapter module. This integration allows the model to leverage the generalization capabilities of large-scale self-supervised learning while maintaining compatibility with the original architecture.

Through a series of experiments on the nuScenes validation set, we quantified the effect of incorporating a foundation model into a 3D detection pipeline. RCDINO achieves a 0.4\% improvement in NDS and 0.7\% in mAP over the baseline, demonstrating the tangible benefit of DINOv2 features for 3D object detection. In a highly competitive benchmark like nuScenes, even fractional gains are crucial, as they reflect meaningful progress toward state-of-the-art performance and improved reliability in real-world applications.



In future work, we plan to further modify the enhanced version of RCTrans by adding a trainable depth estimation module that leverages radar data. This will enable more accurate transformation of image features into the BEV space and has the potential to further improve object detection performance.

\section*{ACKNOWLEDGEMENTS}

The authors would like to thank Vega-GAZ LLC company and the engineering team of the Center for Scientific Programming at MIPT for their assistance in collecting the VegaFull dataset.

\section*{FUNDING}

This research received no specific grant from any funding agency in the public, commercial, or not-for-profit sectors.

\section*{ETHICS APPROVAL AND CONSENT TO PARTICIPATE}

This work does not contain any studies involving human and animal subjects.

\section*{CONFLICT OF INTEREST}

The authors of this work declare that they have no conflicts of interest.

{
\renewcommand{\mkbibbrackets}[1]{#1}
\DeclareFieldFormat{labelnumber}{#1.}
\printbibliography
}

\end{document}